\documentclass[lettersize,journal]{IEEEtran}
\usepackage{amsmath,amsfonts}
\usepackage{algorithmic}
\usepackage{algorithm}
\usepackage{array}
\usepackage[caption=false,font=normalsize,labelfont=sf,textfont=sf]{subfig}
\usepackage{textcomp}
\usepackage{stfloats}
\usepackage{url}
\usepackage{verbatim}
\usepackage{graphicx}
\usepackage{cite}
\usepackage{booktabs}
\usepackage[table]{xcolor}
\usepackage{colortbl}
\usepackage{amsmath}
\usepackage{makecell}
\usepackage{placeins}
\usepackage{xstring}
\usepackage{tabularx}
\usepackage{svg}
\usepackage{multirow}
\usepackage{needspace}
\usepackage{hyperref}
\usepackage[numbers, sort&compress]{natbib}
\newcolumntype{Y}{>{\centering\arraybackslash}X}
\begin{document}

\title{U-STS-LLM: A Unified Spatio-Temporal Steered \\Large Language Model for Traffic Prediction and Imputation}

\author{
  \IEEEauthorblockN{
    Yichen~Zhang,
    Yiyang~Ni,
    Pengcheng~Xia,
    Jun~Li,
    and Chunguo~Li
  }
  \thanks{
    Yichen~Zhang, Jun~Li and Chunguo~Li are with the School of Information Science and Engineering, Southeast University, Nanjing 211189, China (e-mail: 213230744@seu.edu.cn; jun.li@seu.edu.cn; chunguoli@seu.edu.cn).
  }
  \thanks{
    Pengcheng~Xia is with the School of Electronic and Optical Engineering, Nanjing University of Science and Technology, Nanjing 210094, China (e-mail: pengcheng.xia@njust.edu.cn).
  }
  \thanks{
    Yiyang~Ni is with the School of Computer Engineering, Jiangsu Second Normal University, Nanjing 210013, China, and also with the Jiangsu Key Laboratory of Wireless Communications, Nanjing University of Posts and Telecommunications, Nanjing 210003, China (e-mail: niyy@jssnu.edu.cn).
  }
}

\markboth{}%
{Shell \MakeLowercase{\textit{et al.}}: A Sample Article Using IEEEtran.cls for IEEE Journals}

\IEEEpubid{}

\maketitle

\begin{abstract}
Accurate spatio-temporal cellular traffic analysis is fundamental to the reliable operation of modern mobile networks, as it underpins load forecasting for congestion avoidance and missing-data imputation for continuity despite transmission errors. While deeply connected, forecasting and imputation have historically evolved as separate sub-fields. Existing models are typically task-specific, decouple temporal and spatial modeling, and rely on cascaded designs that accumulate errors and latency. Meanwhile, adapting pre-trained LLMs for this domain often leaves spatial structure under-guided and tends to overfit on limited data. To bridge these gaps, we propose U-STS-LLM, a unified framework built on a partially frozen spatio-temporally attentive (PFSTA) LLM backbone. A Dynamic Spatio-Temporal Attention Bias Generator synthesizes a persistent functional graph with transient nodal states to steer attention, while a lightweight parallel branch yields a stable local estimate fused via a Gated Adaptive Fusion, confining the LLM to residual correction. Trained under a unified multi-task objective, the model attains state-of-the-art results with low trainable-parameter overhead. Extensive experiments on real-world cellular traffic datasets demonstrate the effectiveness and transferability of U-STS-LLM, achieving average RMSE improvements of 23.7\% over the strongest baselines in forecasting and 13.1\% in imputation, with further gains exceeding 33\% in zero-shot transfer to unseen network regions.
\end{abstract}


\section{Introduction}
\label{sec:introduction}
\IEEEPARstart{W}{ith} the commencement of large-scale deployment of 5G networks and the massive access of mobile terminals, new network services such as integrated sensing and communication (ISAC) services, virtual reality, and augmented reality are continuously emerging~\cite{Ma2025MobiMixer}. Mobile network operators are not only required to handle the explosive growth of traffic data but are also confronted with the severe challenges posed by its highly complex intrinsic patterns and dynamic evolution~\cite{Zhai2026joint}. Spatio-temporal analysis of cellular traffic yields a characterization of cell load and service distribution, which underpins operational decisions on proactive radio resource allocation, inter-cell load balancing, congestion and admission control, and energy-efficient base station sleeping~\cite{Chen2020adaptive}. In this context, acquiring complete and continuous cellular traffic data and performing accurate traffic forecasting based on it have become critical enabling technologies for ensuring Quality of Service (QoS), optimizing resource efficiency, and reducing operational costs~\cite{Li2026IndustrialTimeSeries, Duan2026cellsyn}. Consequently, the imputation and forecasting of mobile traffic data, particularly the challenging tasks of imputation under high missing rates and long-term forecasting, have garnered extensive attention from both industry and academia~\cite{Guo2025user, Gong2025SpatiotemporalTransformerMobile}. In reality, traffic data generated by massive numbers of base stations and user equipment are often subject to missing values due to failures, transmission errors, and incomplete coverage~\cite{Deng2023graph}, and exhibit characteristics such as high non-stationarity, abrupt variations, and complex spatio-temporal dependencies. Shaped collectively by persistent urban rhythms and dynamic large-scale events, traffic data patterns form intricate global interdependencies, posing serious challenges for traffic imputation and forecasting.

Fundamentally, cellular traffic constitutes time series data. Different variables within a sequence represent information from distinct spatial locations, implying that beyond temporal trends and seasonal patterns, it inherently embodies complex spatial interactions among various cells~\cite{Zhai2026joint, Li2024stgnnm}. The forecasting and imputation of time series have evolved from traditional statistical methods to deep learning techniques based on Recurrent Neural Networks (RNNs)~\cite{Kong2023graphImp}, Convolutional Neural Networks (CNNs)~\cite{Wu2019Graphwavenet}, Graph Neural Networks (GNNs)~\cite{Li2017DCRNN}, Diffusion~\cite{Zheng2024diffusion}, and Transformers~\cite{Li2023LightweightSpatialTemporalTransformer}. However, these models are often tailored to specific tasks and datasets, demanding substantial domain expertise. Meanwhile, constrained by the limited scale and diversity of existing public benchmark datasets, researchers struggle to train deeper models with larger parameter capacities~\cite{Liu2025STLLMplus}. Consequently, the feature representations learned by such models are typically insufficient and not generalizable, as in the related challenge of applying federated learning to heterogeneous and non-IID data~\cite{Ma2022federated, Zhang2023federated, Ma2020safeguarding}. This leads to difficulties in robustly transferring the learned patterns to new network environments or geographical regions.

\IEEEpubidadjcol

In recent years, researchers have begun to explore adapting pre-trained large language models (LLMs) to time-series and spatio-temporal data analysis, leveraging the massive parameters of LLMs to abstract high-level semantic features, while harnessing their inherent robustness and generalization capabilities~\cite{Brown2020language}. The rationale rests on two observations: spatio-temporal sequences exhibit sequential dependencies and intricate long- and short-term contextual associations akin to those in natural language~\cite{Liu2025STLLMplus}, and the target tasks can be uniformly formulated as sequence-to-sequence mappings. For instance, several approaches, including parameter-efficient fine-tuning~\cite{Zhou2023one} (e.g., Low-Rank Adaptation~\cite{Hu2022Lora}), text prompting~\cite{Chen2025LLM}, and reprogramming~\cite{Jin2024TIMELLM}, attempt to align time-series data with the natural-language modality while preserving the pre-trained LLM's capacity for capturing multi-scale complex patterns.

Nevertheless, existing LLM-based spatio-temporal methods still share three structural limitations that prevent the full potential of pre-trained LLMs from being realized in traffic sequence tasks: (1) task decoupling, where models are designed for a single task, leaving transferable knowledge across tasks unshared; (2) serial chaining, where cascaded pipelines induce error propagation, outlier sensitivity, latency accumulation, and potential catastrophic forgetting in LLMs; and (3) spatio-temporal decoupling, where temporal and spatial dependencies are modeled in separate branches, such that the inherent coupling between them remains under-explored.

To address the three limitations above, we propose U-STS-LLM, a unified spatio-temporal steered LLM. The framework tackles traffic forecasting and imputation within a single architecture. At its core lies a partially frozen spatio-temporally attentive (PFSTA) pre-trained LLM backbone. We inject a spatio-temporal attention bias into this backbone to encode transferable dependencies, which enables joint spatial–temporal modeling in a parameter-efficient regime. In parallel, a lightweight dedicated branch runs alongside the LLM and supplies learnable signals embedding task-specific inductive biases. Extensive experiments on real-world cellular traffic datasets demonstrate that U-STS-LLM achieves state-of-the-art performance, with average MAE and RMSE improvements of 23.9\% and 23.7\% for long-horizon forecasting and 11.5\% and 13.1\% for high-missing-rate imputation, while maintaining high efficiency and strong zero-shot generalization.

Our specific contributions are as follows.

\begin{itemize}
\item We frame the problem under a unified multi-task optimization objective, jointly training the model for long-term forecasting and imputation under high missing rates. By learning cross-task shared patterns and transferring knowledge between tasks, the model acquires a unified representation of the underlying spatio-temporal process. This compels it to develop a holistic understanding of the spatio-temporal data manifold, enhancing the robustness and generalization of both tasks.

\item To the best of our knowledge, this work presents the first LLM-parallel architecture tailored to traffic data. Specifically, a lightweight branch runs in parallel with the LLM to supply a robust local estimate, and a Gated Adaptive Fusion module fuses this estimate with the LLM-refined output. This design formulates learning as residual correction, steering the LLM toward deeper inference while preserving its pretrained reasoning ability and generalization capacity. It thereby mitigates overfitting to macro-trends and alleviates the alignment dilemma common in fine-tuning large foundation models on small structured datasets, leading to improved accuracy.

\item We design a dynamic spatio-temporal attention bias generator that constructs priors from historical traffic-profile similarity and transient nodal states. The resulting spatio-temporal attention map is injected as a bias into the attention mechanism of the LoRA-tuned LLM backbone. This joint modeling of spatio-temporal heterogeneity requires no costly physical spatial knowledge and ensures portability across network deployments.

\end{itemize}

The remainder of this paper is organized as follows. Section II reviews related work on temporal forecasting/imputation and LLMs for spatio-temporal series. Section III details the proposed U-STS-LLM framework. Section IV presents the experimental setup, results, and analysis. Finally, Section V concludes the paper.


\section{Related Works}
\label{sec:related_works}

\subsection{Temporal Forecasting and Imputation}
 Forecasting involves predicting the features of future time steps based on data from historical time steps, whereas imputation aims to estimate missing values in one or more time series. The more challenging regimes, however, are long-horizon forecasting and imputation under high missing rates: multi-step forecasting progressively accumulates errors across successive horizons, while severe missingness markedly weakens the exploitable spatio-temporal autocorrelation that imputation relies upon. Traditional time series analysis tools, such as Support Vector Regression (SVR)~\cite{Cao2003SVR}, Vector AutoRegression (VAR)~\cite{Biller2003VAR}, and AutoRegressive Integrated Moving Average (ARIMA)~\cite{Pierce1970ARIMA}, as well as specialized traditional statistical methods and machine learning methods in the imputation field, like linear interpolation~\cite{Kuffel1997Linear}, EM algorithms~\cite{Moon1996EM}, and K-nearest neighbors~\cite{Samet2008KNearest}, struggle to handle complex temporal dependencies and non-linear relationships, leading to lower accuracy in results~\cite{Jin2024SurveyGNN}.

With the development of deep learning techniques, Long Short-Term Memory (LSTM)~\cite{Hochreiter1997LSTM} and its more efficient simplified version, Gated Recurrent Unit (GRU)~\cite{Cho2014GRU}, endowed RNNs with the capability to handle long sequence data by introducing gating mechanisms. BRITS~\cite{Cao2018BRITS}, building upon LSTM, proposed an end-to-end bidirectional recurrent imputation framework. Temporal Convolutional Network (TCN)~\cite{Bai2018TCN} and Transformer~\cite{Zhao2026TransformerSurvey} further enhanced models' ability to identify long-term temporal dependencies. SAITS~\cite{Du2023SAITS} introduced the self-attention mechanism and cross-feature correlations into the time series imputation domain. TimesNet~\cite{Wu2023TimesNet} stands out among them; by incorporating a frequency-domain perspective, it achieved adaptive capture of multi-periodicity, delivering consistent state-of-the-art performance across multiple mainstream time series tasks, including forecasting and imputation. Although these models perform reasonably well on their designated tasks, they do not explicitly model cross-node spatial dependencies, and their accuracy degrades under non-stationary conditions.

Subsequently, Spatial-Temporal Graph Convolutional Network (STGCN)~\cite{Yu2018STGCN} approximated graph convolution using Chebyshev polynomials, interleaving them with temporal convolutions to capture both spatial and temporal patterns simultaneously. Subsequent models like StemGNN~\cite{Cao2020StemGNN}, DCRNN~\cite{Li2017DCRNN}, Graph WaveNet~\cite{Wu2019Graphwavenet}, STDGDC~\cite{Guo2025user}, and the second version of STGCN~\cite{Yu2018STGCN} strengthened the modeling and learning capacity of spatial dependencies by introducing mechanisms such as spectral methods, message passing, graph diffusion, GAT, and GCN. However, these methods decouple the temporal and spatial dimensions. They also place high demands on computational resources, exhibit limited robustness against adversarial perturbations and distribution shifts, struggle to adapt to dynamic systems~\cite{Jin2024SurveyGNN}, and suffer from insufficient generalization capability.

\subsection{Large Language Models for Spatio-Temporal Series}
The adaptation of LLMs to sequential data has extended from natural language into the realm of numerical time series analysis. This is motivated by the shared sequential nature of both data types and the LLMs' proven capacity for modeling long-range dependencies~\cite{Liang2024foundation}. Initial explorations primarily targeted temporal forecasting, employing strategies like fine-tuning~\cite{Zhou2023one}, reprogramming~\cite{Jin2024TIMELLM}, and prompt-based learning~\cite{Chen2025LLM}. Notably, works like Voice2Series~\cite{Yang2021voice2series} and LLMTime~\cite{Gruver2023LLMTime} demonstrated that knowledge from linguistic models could be transferred to time series tasks. Following this, recent methods like NPP-GPT~\cite{Chang2026NPPGPT} have further refined these strategies for domain-specific forecasting, emphasizing parameter-efficient adaptation and explicit modality alignment. However, these methods are typically specialized for either forecasting or imputation, leaving transferable knowledge across tasks unshared. Meanwhile, the aforementioned approaches predominantly focus on isolated temporal patterns, overlooking the spatial correlations within cellular networks and thus yielding suboptimal accuracy.

To incorporate spatial awareness, existing studies predominantly follow two methodological paradigms. However, both rely uniformly on purely serial architectures, which can lead to error accumulation and increased latency, and render the LLM more prone to overfitting during fine-tuning. Moreover, they model spatial relations in isolation, failing to capture the spatio-temporal dependencies and heterogeneity across base stations.

The first paradigm treats spatial elements as discrete tokens, directly feeding the serialized time series of each node into the large language model. Models like LLM‑TFP~\cite{Cheng2025TFP} and ST‑LLM~\cite{Liu2024STLLM} employ methods like token generators to produce token embeddings, relying on the self-attention mechanism to implicitly capture relationships between nodes. While this approach offers high modeling flexibility, it sacrifices explicit structural guidance. Due to the model's lack of prior knowledge about node interactions, it can lead to training instability and inefficient learning of spatial dependencies with limited data. A variant of this paradigm primarily uses the large language model as a semantic knowledge enhancer. Frameworks such as STCInterLLM~\cite{Li2025STCInterLLM}, UrbanGPT~\cite{Li2024UrbanGPT}, and GraphGPT~\cite{Tang2024Graphgpt} inject spatial contextual information into the large language model via graph-based encoders. In practice, this often prevents the LLM's sequential reasoning capabilities from being directly applied to spatio-temporal modeling, potentially underutilizing its capacity and complicating the training pipeline.

The second, more integrative paradigm aims to explicitly combine the large language model with graph-structured modules. For instance, GATGPT~\cite{Chen2023GATGPT} and ST‑LLM+~\cite{Liu2025STLLMplus} integrate token embeddings with Graph Attention Networks to explicitly handle spatial relationships; TAGSC~\cite{Li2026graph} utilizes Graph Convolutional Networks to extract spatial features. The core idea is to enhance the large language model with a learnable prior of graph-structured relationships. Yet these methods typically only generate a task-specific, prior-free embedding for each node, or rely on static graphs constructed from physical spatial adjacency, which can be expensive to obtain. This can limit their generalizability to new areas with different topological structures, and their task-specific designs may result in insufficiently comprehensive representations.

\needspace{2\baselineskip}

\section{Proposed Framework}
\label{sec:framework}

\subsection{Overview}
\label{subsec:overview}
Forecasting future states and reconstructing missing values in spatio-temporal traffic data, 
cellular network loads for example, are intrinsically linked challenges that demand models capable of capturing complex, long-range dependencies amidst pervasive data incompleteness. To this end, we present the Unified Spatio-Temporal Steered Large Language Model (U-STS-LLM). The model is architected as a multi-task framework that processes a spatio-temporal input sequence \( \mathbf{X} \in \mathbb{R}^{B \times L \times N \times C} \) and produces a corresponding output \( \mathbf{Y} \in \mathbb{R}^{B \times S \times N \times C} \), with \(B,L,N,C\) denoting batch size, input sequence length, number of spatial nodes, and feature dimension, respectively, and \(S\) denoting the target sequence length (e.g., the forecasting horizon or \(L\) for imputation).

U-STS-LLM combines a partially frozen spatio-temporally attentive (PFSTA) LLM backbone with explicit domain-specific inductive biases and a lightweight parallel branch. Initially, the raw input undergoes conditioning to extract fundamental periodic patterns and a distilled global spatio-temporal signature. In parallel, a dynamic spatio-temporal attention bias synthesizes a persistent functional graph with transient nodal states from the input's structure, serving as a soft prior for relational reasoning. Subsequently, the embedded features are processed by the core PFSTA backbone, a GPT-2 model adapted via LoRA. The dynamic bias is injected into the attention mechanism of the top layers, steering the LLM's self-attention toward both persistent graph-topological relationships and transient nodal-state correlations inherent in traffic systems. Finally, the deep representations from the backbone are combined with a robust local estimate from the parallel branch via a Gated Adaptive Fusion module, confining the LLM to residual correction and yielding the refined final output. The proposed architecture follows a complementary design: the LLM provides the capacity for modeling long-term temporal dependencies and complex interactions, while the injected bias and guidance signal help direct these capacities toward the structural and statistical regularities of the target domain, allowing a single model to handle both prediction and imputation.

\begin{figure*}[t]
    \centering
    \includegraphics[width=1\linewidth]{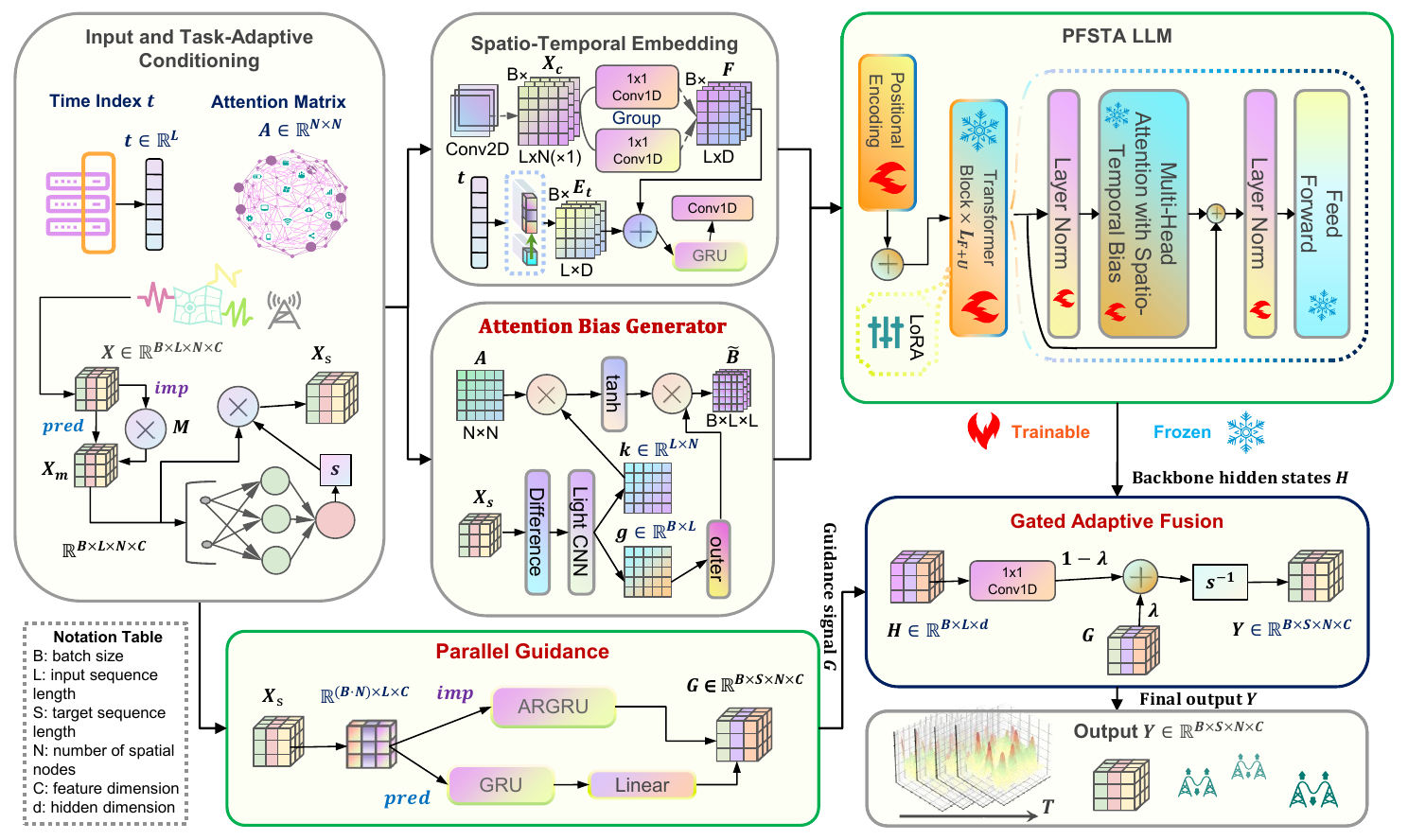}
    \caption{The overall architecture of the proposed U-STS-LLM framework. The U-STS-LLM framework first conditions the input via dynamic scaling. Two parallel pathways then process it: one distills a global spatio-temporal context signal, while the other generates a stable, node-wise preliminary estimate. A dynamic graph-based bias, synthesized from the input's structure, is generated to steer relational reasoning. This bias is injected into the attention mechanism of the top layers in the Partially Frozen Spatio-Temporal Attention (PFSTA) backbone, a LoRA-adapted LLM whose lower layers remain frozen. The deep, steered representations from the PFSTA are finally fused with the preliminary estimate to produce the refined output for unified forecasting and imputation.}
    \label{fig:placeholder}
\end{figure*}

\subsection{Input and Task-Adaptive Conditioning}
\label{subsec:input}
The initial processing stage is task-specific. For imputation, we simulate realistic block missingness patterns by applying a stochastic binary mask \( \mathbf{M} \in \{0,1\}^{B \times L \times N \times C} \), where each element is independently masked (set to 0) with a probability \(p\) sampled uniformly from a predefined range \([r_{\text{min}}, r_{\text{max}}]\). The masked input is then \( \mathbf{X}_m = \mathbf{X} \odot \mathbf{M} \). For prediction, the input is used directly, i.e., \( \mathbf{X}_m = \mathbf{X} \).

Real-world traffic data is inherently non-stationary and exhibits scale variations across nodes, time periods, and systems. To counter such distributional shifts, we propose a Conditional Input Scaling mechanism that computes a dynamic scaling factor from input statistics, rather than applying fixed normalization. A lightweight, fully-connected network \(g_{\phi}\) maps a global statistic \( \psi(\mathbf{X}) \) (e.g., the mean feature value across the last observed time step) to the perturbation bounds of a uniform distribution:
\begin{equation}
\begin{aligned}
[\Delta a, \Delta b] = g_{\phi}(\psi(\mathbf{X})), \\
\quad \tilde{a}_{\mathbf{X}} = c + \Delta a \cdot \delta, \quad \tilde{b}_{\mathbf{X}} = c + \Delta b \cdot \delta,  \\
\quad a_{\mathbf{X}} = \min(\tilde{a}_{\mathbf{X}}, \tilde{b}_{\mathbf{X}}), 
\quad b_{\mathbf{X}} = \max(\tilde{a}_{\mathbf{X}}, \tilde{b}_{\mathbf{X}}),
\end{aligned}
\label{eq:cond_scale}
\end{equation}
where \(c\) is an iteratively updated base center, and \(\delta\) is a learnable parameter representing the fluctuation radius. A scale factor \( s \sim \mathcal{U}(a_{\mathbf{X}}, b_{\mathbf{X}}) \) is then sampled via the reparameterization trick, and applied to the input: \( \mathbf{X}_s = s \cdot \mathbf{X}_m \). Crucially, the inverse scaling $s^{-1}$ is applied to the final model output before loss computation. This conditions the model's activations on input magnitude, thereby encouraging scale-invariant representations and improving generalization to unseen data regimes.

\subsection{Spatio-Temporal Feature Embedding and Fusion}
\label{subsec:embed}
This component produces two complementary information streams: a low-dimensional global context and a node-specific preliminary estimate, capturing both macroscopic trends and fine-grained local patterns. By mapping deterministic time indices to learnable embeddings, the Temporal Embedding module provides strong inductive biases for periodic patterns. For a time step \(t\), we define its intra-day index \(h_d(t) = \lfloor t / f \rfloor \mod Q\) and its intra-week index \(h_w(t) = \lfloor t / (Q \cdot f) \rfloor \mod P\), where \(f\) is the time resolution factor, and \(Q\) and \(P\) are the number of intervals per day and week, respectively. The combined temporal embedding is obtained via:
\begin{equation}
\mathbf{E}_t = \mathbf{W}_d[h_d(t)] + \mathbf{W}_w[h_w(t)] \in \mathbb{R}^{B \times L \times D},
\label{eq:temp_emb}
\end{equation}
where \(\mathbf{W}_d \in \mathbb{R}^{Q \times D}\) and \(\mathbf{W}_w \in \mathbb{R}^{P \times D}\) are learnable embedding matrices. This formulation encodes cyclical temporal prior knowledge, facilitating the learning of recurring patterns such as daily and weekly rhythms inherent in traffic data.

The Spatio-Temporal Perception pathway distills the high-dimensional input into a compact, global representation. First, the channel dimension is compressed using a two-dimensional convolution:
\begin{equation}
\mathbf{X}_c = \text{GELU}\left(\text{Conv2D}_{1\times1}(\mathbf{X}_s)\right) \in \mathbb{R}^{B \times L \times N}.
\label{eq:channel_compress}
\end{equation}
Subsequently, a one-dimensional grouped convolution is applied over the spatial dimension, partitioning the \(N\) nodes into \(D\) groups to extract shared temporal dynamics within each group:
\begin{equation}
\mathbf{F} = \text{GELU}\left(\text{GroupConv1D}(\mathbf{X}_c)\right) \in \mathbb{R}^{B \times L \times D}.
\label{eq:group_conv}
\end{equation}
The group-level feature \(\mathbf{F}\) is then fused with the temporal embedding \(\mathbf{E}_t\) and further refined by a Gated Recurrent Unit (GRU) to capture sequential dependencies:
\begin{equation}
\mathbf{H}_g = \text{GRU}\left(\mathbf{F} + \tanh(\mathbf{E}_t)\right) \in \mathbb{R}^{B \times T \times d_h},
\label{eq:gru_feat}
\end{equation}
where \(d_h\) is the hidden dimension of the GRU. Finally, a convolutional layer transforms this representation into the model's unified hidden space as
$\mathbf{H}_p = \text{Conv1D}(\mathbf{H}_g) \in \mathbb{R}^{B \times L \times d}$.
This pathway abstracts a compact global spatio-temporal context \(\mathbf{H}_p\) as input to the LLM backbone, reducing computational cost while retaining essential information.

Concurrently, the Guidance Signal Generation pathway produces a stable, node-wise preliminary estimate. The scaled input $\mathbf{X}_s$ is first reshaped to treat each node's feature sequence independently, resulting in a tensor $\mathbf{X}_r \in \mathbb{R}^{(B \cdot N) \times L \times C}$. A task-specific Gated Recurrent Unit (GRU) network then processes these sequences. For the prediction task, a unidirectional GRU is employed. For the imputation task, a specialized bi-directional Auto-Regressive GRU (ARGRU) is utilized. The key design of the ARGRU is to explicitly model the imputation estimate at each time step as a variable within the computational graph, which is then refined using the contextual information from both directions. This process is conditioned on the input mask $\mathbf{M}_r$: for a missing value at step $t$, the estimate $\hat{\mathbf{x}}_t$ is computed from the hidden states, and the update follows $\tilde{\mathbf{x}}_t = \mathbf{m}_t \odot \mathbf{x}_t + (1-\mathbf{m}_t) \odot \hat{\mathbf{x}}_t$. By refining each missing value with both past and future context across the entire sequence, this mask-conditioned bidirectional design compensates for the causal-mask limitation of pre-trained LLMs, thereby turning the preliminary reconstruction into a learnable, context-aware signal. The process is formalized as:
\begin{equation}
    \mathbf{G}_r = 
    \begin{cases}
        \text{Linear}\bigl(\mathbf{GRU}(\mathbf{X}_r)\bigr),      & \text{for prediction}, \\
        \mathbf{ARGRU}(\mathbf{X}_r), & \text{for imputation},
    \end{cases}
    \label{eq:guide_gru}
\end{equation}
where the output tensor is of dimension $\mathbb{R}^{(B \cdot N) \times L \times C}$.

The output is then reshaped back to the original spatio-temporal format. For the prediction task, only the final \(S\) time steps are retained to align with the forecast horizon, yielding the guidance signal \(\mathbf{G} \in \mathbb{R}^{B \times S \times N \times C}\), whereas for imputation, the full sequence of length \(L\) is used. This guidance signal \(\mathbf{G}\) acts as a robust, locally-consistent baseline. Its generation through a simple per-node recurrent model ensures computational efficiency and provides a direct learning pathway. The subsequent fusion with the LLM output enables residual learning, in which the LLM refines the baseline by modeling cross-node interactions and long-range dependencies that the local model may not adequately capture.

\subsection{Dynamic Spatio-Temporal Attention Bias Generator}
\label{subsec:bias_gen}
Static graph adjacency matrices, while encoding topological proximity, fail to represent the time-varying intensity of functional influence between nodes. To construct an adaptive relational prior, we propose a generator for a Dynamic Spatio-Temporal Attention Bias matrix \( \tilde{\mathbf{B}} \in \mathbb{R}^{B \times L \times L} \). The process begins by computing a normalized differential signal \( \mathbf{S} \in \mathbb{R}^{B \times L \times C} \) from \( \mathbf{X}_s \) to highlight regions of change. A lightweight convolutional network \(f_{\text{gate}}\), composed of a few 1D convolutional layers with GELU activations, then processes \( \mathbf{S} \) to produce two components: a temporal modulation matrix \( \mathbf{k} \in \mathbb{R}^{L \times N} \) and a dynamic factor matrix \( \mathbf{g} \in \mathbb{R}^{B \times L} \).

A static communication adjacency matrix \( \mathbf{A} \in \mathbb{R}^{N \times N} \) is pre-computed once from the training dataset, typically based on the pairwise similarity (e.g., dot product similarity) of nodes' historical profiles, thereby encoding persistent, long-term functional correlations. A learnable projection matrix \( \mathbf{P} \in \mathbb{R}^{L \times N} \), modulated by \( \mathbf{k} \), contracts this spatial graph into a temporal bias prior:
\begin{equation}
\mathbf{P}' = \tanh(\mathbf{P} \odot \text {k}), \quad \mathbf{B}_s = \mathbf{P}' \mathbf{A} \mathbf{P}'^{\top} \in \mathbb{R}^{L \times L}.
\label{eq:static_bias}
\end{equation}
The final dynamic bias is computed as the Hadamard product of this static structural bias and an outer product derived from the instantaneous dynamic factors, scaled by a learnable intensity parameter \( \mu \):
\begin{equation}
\tilde{\mathbf{B}} = \mu \cdot (\mathbf{g} \mathbf{g}^{\top}) \odot \mathbf{B}_s.
\label{eq:dyn_bias}
\end{equation}
This formulation is compositional: \( \mathbf{B}_s \) provides a stable template of potentially related time-step pairs based on the pre-defined graph structure projected through \( \mathbf{P}' \), while the rank-1 matrix \( \mathbf{g} \mathbf{g}^{\top} \) dynamically modulates the strength of these potential connections according to the current input. The parameter \( \mu \) controls the overall strength of this inductive bias. This allows the attention mechanism to be continuously steered by both enduring spatial relationships and transient nodal activities.

\subsection{Partially Frozen Spatio-temporal Attention (PFSTA) with LoRA Tuning}
\label{subsec:pfsta}
The core sequence modeling capability is furnished by a pre-trained GPT-2 Transformer, utilized as a universal sequence processor. We adopt a partially frozen strategy to balance knowledge retention from pre-training with task-specific adaptation. The lower \(L_f\) transformer blocks are kept entirely frozen to preserve the rich, general-world sequence modeling knowledge. For the top \(U\) blocks, we selectively unfreeze parameters to enable adaptation. The operations within the \(l\)-th transformer block are summarized as follows.
\begin{equation}
\begin{aligned}
\mathbf{H}' &= \text{LayerNorm}(\mathbf{H}^{l-1}), \\
\mathbf{Z} &= \text{MHA}(\mathbf{H}', \mathbf{H}', \mathbf{H}') + \mathbf{H}^{l-1}, \\
\mathbf{H}^l &= \text{FFN}(\text{LayerNorm}(\mathbf{Z})) + \mathbf{Z},
\end{aligned}
\label{eq:transformer_block}
\end{equation}
where MHA denotes the Multi-Head Attention module and FFN the Feed-Forward Network. In our PFSTA strategy, for the top \(U\) blocks, the parameters of the MHA modules are made trainable, while those of the FFN modules remain frozen. This largely confines trainable adaptation to the attention modules, while the frozen FFNs serve as fixed nonlinear processors. As a result, the risk of overfitting to the limited target-domain data is reduced.

Parameter-efficient fine-tuning is achieved by integrating LoRA into the trainable MHA modules. For a frozen pre-trained weight matrix \( \mathbf{W}_0 \in \mathbb{R}^{d \times d} \) within the attention projections (e.g., for query, key), its update is constrained to a low-rank decomposition as
\begin{equation}
\mathbf{W} = \mathbf{W}_0 + \Delta \mathbf{W} = \mathbf{W}_0 + \mathbf{B} \mathbf{A},
\label{eq:lora}
\end{equation}
where \( \mathbf{B} \in \mathbb{R}^{d \times r}, \mathbf{A} \in \mathbb{R}^{r \times d} \), and the rank \( r \ll d \). Only the parameters in \( \mathbf{B} \) and \( \mathbf{A} \) are updated during training, reducing the number of trainable parameters for that weight matrix from \(O(d^2)\) to \(O(2dr)\), which is crucial for efficiently fine-tuning large foundation models.

Graph-informed attention steering is implemented by injecting the dynamic bias \( \tilde{\mathbf{B}} \) into the attention computation of the top \(U\) blocks. The perception feature \( \mathbf{H}_p \) serves as the input sequence to the PFSTA backbone. The attention scores are thus computed as
\begin{equation}
\text{Attention}(\mathbf{Q}, \mathbf{K}, \mathbf{V}, \tilde{\mathbf{B}}) = \text{Softmax}\left( \frac{\mathbf{Q}\mathbf{K}^{\top}}{\sqrt{d_k}} + \tilde{\mathbf{B}} \right) \mathbf{V},
\label{eq:attn_with_bias}
\end{equation}
where \( \mathbf{Q} = \mathbf{H}_p \mathbf{W}_q, \mathbf{K} = \mathbf{H}_p \mathbf{W}_k \). The additive bias \( \tilde{\mathbf{B}} \) directly and adaptively adjusts the pre-softmax attention logits between any pair of time steps \((i, j)\). High values in \( \tilde{\mathbf{B}}_{ij} \) encourage the model to attend more strongly from step \(i\) to step \(j\), as informed by the compound spatio-temporal prior in \eqref{eq:dyn_bias}. This mechanism steers the LLM's content-based attention toward connections that are known to be meaningful in the traffic domain, without modifying the core Transformer architecture. The frozen lower blocks ensure the model retains its fundamental sequence understanding, while the augmented top layers specialize in domain-relevant relational reasoning.

\subsection{Gated Adaptive Fusion for Output Refinement}
\label{subsec:fusion}
Let \(\mathbf{H}_{\text{pfsta}} \in \mathbb{R}^{B \times L \times d}\) be the output hidden
states of the PFSTA backbone. \(\mathbf{H}_{\text{pfsta}}\) is then projected to the target feature space via a task-specific regression convolutional layer and reshaped to match the output dimensions, i.e.,
\begin{equation}
\mathbf{H}_{\text{out}} = \text{Reshape}(\text{RConv} (\mathbf{H}_{\text{pfsta}})) \in \mathbb{R}^{B \times S \times N \times C}.
\label{eq:regression}
\end{equation}

The final model output is synthesized via a Gated Adaptive Fusion of this LLM-refined representation and the previously generated guidance signal as
\begin{equation}
\begin{aligned}
\mathbf{Y}_{\text{out}} = (1 - \lambda) \cdot \mathbf{H}_{\text{out}} + \lambda \cdot \mathbf{G}, \\
\mathbf{Y} = {s^{-1}}\cdot \mathbf{Y}_{\text{out}},
\label{eq:fusion}
\end{aligned}
\end{equation}
where \( \lambda \in (0, 1) \) is a trainable gating scalar. This fusion eases training by providing a task-specific baseline and frames the learning objective as predicting residual corrections relative to that baseline.

\subsection{Multi-Task Optimization Objective}
\label{subsec:loss}
The U-STS-LLM framework is optimized end-to-end with a composite loss function that jointly supervises the prediction and imputation tasks, i.e., 
\begin{equation}
\mathcal{L}_{\text{total}} = \alpha \cdot \mathcal{L}_{\text{pred}} + (1 - \alpha) \cdot \mathcal{L}_{\text{imp}},
\label{eq:total_loss}
\end{equation}
where \( \alpha \in [0,1] \) is a hyperparameter balancing the two objectives. The prediction loss \( \mathcal{L}_{\text{pred}} \) is the standard Mean Squared Error (MSE) between the forecasted sequence \( \mathbf{Y} \) and the ground-truth future values \( \mathbf{Y}_{\text{true}} \) over all nodes and features. The imputation loss \( \mathcal{L}_{\text{imp}} \) is also MSE, but it is computed exclusively over the elements that were actively masked by the binary mask \( \mathbf{M} \). Formally, we have
\begin{equation}
\mathcal{L}_{\text{imp}} = \frac{1}{|\Omega|}\sum_{(b,l,n,c) \in \Omega} \left( \mathbf{Y}_{b,l,n,c} - \mathbf{X}_{b,l,n,c} \right)^2,
\label{eq:imp_loss}
\end{equation}
where \( \Omega = \{(b,l,n,c) | \mathbf{M}_{b,l,n,c} = 0 \} \) is the set of masked indices. This restricted computation forces the model to learn the underlying data distribution for generating plausible values at unobserved spatio-temporal locations, rather than trivial memorization or identity mapping of the visible inputs. The joint optimization under this combined loss encourages the shared parameters to learn a unified representation of the data's spatio-temporal dynamics, which supports both extrapolation for forecasting and interpolation for imputation.

\subsection{Complexity Analysis}
\label{subsec:complexity}
The computational complexity of U-STS-LLM is primarily dictated by the self-attention mechanism in its PFSTA backbone, though the specific profile is shaped by the model's design. Here \(L\), \(N\), \(d\), and \(r\) denote the sequence length, number of spatial nodes, hidden dimension of the LLM backbone, and LoRA rank, respectively. The time complexity arises from two main parts of the backbone: the frozen lower layers, which perform standard self-attention at a cost of \(\mathcal{O}(L^2 d)\), and the unfrozen top layers that employ graph-steered attention. Generating the dynamic bias \(\tilde{\mathbf{B}}\) for the top layers incurs an additional \(\mathcal{O}(L N^2 + L^2 N)\) cost. The Spatio-Temporal Perception and Guidance Generation modules, which rely on convolutions and GRUs, scale linearly with sequence length and node count, contributing only minor overhead. The model achieves parameter efficiency through partial freezing and LoRA. Although the total parameter count is dominated by the frozen LLM backbone, the number of trainable parameters in each projection weight is reduced to \(\mathcal{O}(2dr)\) instead of \(\mathcal{O}(d^2)\). In terms of memory, the model requires \(\mathcal{O}(L^2)\) for attention weights and \(\mathcal{O}(L d)\) for layer activations. The partial freezing strategy further reduces peak memory usage during training, as intermediate activations do not need to be stored for the frozen layers. Thus, U-STS-LLM leverages a large pre-trained model while remaining efficient for training and inference on spatio-temporal data.

\section{Experiments and Implementation Details}
\label{sec:experiments}

\subsection{Datasets}
\label{subsec:datasets}
The data utilized in this study originate from the Milan telecommunications dataset~\cite{DVN/EGZHFV_2015} of the ``Telecom Italia Big Data Challenge,'' containing anonymized telecommunication records for Milan from November 1, 2013, to January 1, 2014. Temporally, the data is aggregated at 10-minute intervals. Spatially, it is standardized onto a city-wide grid of 10,000 (100x100) squares, with each square approximately 235 meters on a side. The dataset specifically records the following telecommunication activities: SMS-in/out activity, Call-in/out activity, and Internet traffic activity. Internet connection records are generated only when a session exceeds 15 minutes in duration or 5 MB in data volume. To preserve privacy, all publicly released interaction count data has undergone uniform proportional scaling for anonymization.

The format of the Trento telecommunications dataset~\cite{DVN/QLCABU_2015} used in the zero-shot experiment is similar to that of the aforementioned dataset, but it covers telecommunications data from 11,466 areas within the province of Trento.

\subsection{Experimental Settings}
\label{subsec:settings}
In the data preprocessing stage, we employed the K-means clustering algorithm to aggregate all nodes into 200 clusters based on spatial proximity. Furthermore, we ensured that nodes which were close in spatial terms were also assigned adjacent index numbers after aggregation. This design offers a dual benefit: it alleviates the memory burden on the GPU and facilitates more efficient model training. To rigorously evaluate the model's generalization capability, we chronologically split the dataset into training, validation, and test sets at a ratio of 8:1:1. Samples were loaded using a sliding window approach, and all data underwent median normalization. For the minimal number of missing values present in the datasets, we applied linear interpolation for imputation. These imputed values were subsequently excluded during the calculation of evaluation metrics to ensure the reliability of our results. 

The experiments were conducted on hardware equipped with an NVIDIA RTX A6000 GPU. The software environment consisted of Python 3.8.19, PyTorch 2.4.1, and CUDA 11.8. For model training, we utilized the AdamW optimizer paired with a learning rate scheduler that combined a linear warm-up phase with a cosine annealing strategy. Early stopping was also introduced to prevent overfitting. The historical time step length was set to $L=288$ (48 hours), and the future time step length to $S=144$(24 hours). For the imputation task, a random masking rate between 0.70 and 0.80 was applied. The primary evaluation metrics adopted in this study were the Mean Absolute Error (MAE) and the Root Mean Square Error (RMSE).

For each comparative experiment, all baseline models (including our proposed U-STS-LLM) followed exactly the same data preprocessing and evaluation pipeline. This includes, but is not limited to, identical batch size, spatial clustering, historical length ($L$), prediction length ($S$), masking rate, early stopping configuration, and evaluation metrics (MAE and RMSE). For almost all baseline models, we utilized their official open-source implementations, making only minimal modifications to accommodate differences in data tensor formats. To further ensure experimental fairness, the LLM component of our framework adopts GPT-2, which belongs to the same generation of pre-trained language models as the backbones used in most LLM-based baselines.

\subsection{Baselines}
\label{subsec:baselines}
We compare U-STS-LLM against a broad range of advanced baselines representing different technical lineages in spatio-temporal analysis. The selected baselines are grouped into several prominent categories to establish a formidable benchmark. First, we include foundational deep learning models: the seminal LSTM~\cite{Hochreiter1997LSTM} for capturing temporal dependencies and its spatial extension ConvLSTM. Second, we consider recent, leading time-series foundation models: the simple yet powerful DLinear~\cite{Zeng2023DLinear} and the state-of-the-art TimesNet~\cite{Wu2023TimesNet}, which redefines temporal modeling by transforming 1D series into 2D space to capture multi-periodicity variations and has set new benchmarks in numerous forecasting tasks. Third, dedicated spatio-temporal architectures are represented by the influential STGCN~\cite{Yu2018STGCN}. Fourth, to benchmark against the most recent paradigm of leveraging LLMs, we include a suite of adaptations: GPT4ST, GATGPT~\cite{Chen2023GATGPT}, GCNGPT, ST-LLM~\cite{Liu2024STLLM}, and its enhanced variant ST-LLM+~\cite{Liu2025STLLMplus}. Finally, for the imputation task, we compare against benchmark deep learning imputation methods, BRITS~\cite{Cao2018BRITS} and SAITS~\cite{Du2023SAITS}, which are recognized for their robust performance in handling missing data.

\begin{table*}[!tbp]
\vspace*{-\baselineskip}
\centering
\caption[PREDICTION PERFORMANCE COMPARISON]{%
  \textbf{PREDICTION PERFORMANCE COMPARISON OF DIFFERENT METHODS \\IN TERMS OF MAE AND RMSE ON NORMALIZED DATA}
}
\label{tab:baseline_prediction}
\resizebox{\textwidth}{!}{%
\begin{tabular}{ccccccccccc}
\toprule
 & \multicolumn{2}{c}{SMS-IN} & \multicolumn{2}{c}{SMS-OUT} & \multicolumn{2}{c}{CALL-IN} & \multicolumn{2}{c}{CALL-OUT} & \multicolumn{2}{c}{INTERNET} \\
\cmidrule(lr){2-3} \cmidrule(lr){4-5} \cmidrule(lr){6-7} \cmidrule(lr){8-9} \cmidrule(lr){10-11}
Model & MAE & RMSE & MAE & RMSE & MAE & RMSE & MAE & RMSE & MAE & RMSE \\
\midrule
{\small LSTM} & 0.2077 & 0.4182 & 0.2873 & 0.5388 & \textcolor{blue}{\underline{0.2094}} & \textcolor{blue}{\underline{0.3991}} & 0.2284 & \textcolor{blue}{\underline{0.3970}} & 0.2116 & 0.4047 \\
{\small ConvLSTM} & 0.2225 & 0.4830 & 0.2821 & 0.5466 & 0.2329 & 0.4725 & 0.2539 & 0.4949 & 0.1831 & 0.3429 \\
{\small DLinear} & 0.5457 & 0.6929 & 0.6172 & 0.7986 & 0.5596 & 0.7445 & 0.5609 & 0.7221 & 0.5847 & 0.7017 \\
{\small Timesnet} & \textcolor{blue}{\underline{0.2006}} & \textcolor{blue}{\underline{0.3825}} & \textcolor{blue}{\underline{0.2654}} & \textcolor{blue}{\underline{0.5057}} & 0.2195 & 0.4284 & 0.2245 & 0.4116 & \textcolor{blue}{\underline{0.1669}} & \textcolor{blue}{\underline{0.2711}} \\
{\small STGCN} & 0.2395 & 0.4906 & 0.3088 & 0.5617 & 0.2391 & 0.4601 & 0.2657 & 0.4866 & 0.2464 & 0.4710 \\
{\small GATGPT} & 0.2592 & 0.5059 & 0.3495 & 0.6204 & 0.2610 & 0.4913 & 0.2731 & 0.4911 & 0.3524 & 0.5533 \\
{\small GPT4ST} & 0.2144 & 0.4370 & 0.3014 & 0.5743 & 0.2145 & 0.4293 & \textcolor{blue}{\underline{0.2228}} & 0.4286 & 0.3089 & 0.4999 \\
{\small GCNGPT} & 0.2578 & 0.5124 & 0.3430 & 0.6200 & 0.2585 & 0.4941 & 0.2701 & 0.4973 & 0.3491 & 0.5534 \\
{\small ST-LLM} & 0.2158 & 0.4637 & 0.2903 & 0.5639 & 0.2258 & 0.4670 & 0.2413 & 0.4757 & 0.3041 & 0.4978 \\
{\small ST-LLM+} & 0.2230 & 0.4557 & 0.2864 & 0.5551 & 0.2358 & 0.4730 & 0.2542 & 0.4902 & 0.2827 & 0.4573 \\
\textbf{\textcolor{red}{U-STS-LLM}} & \textbf{\textcolor{red}{0.1409}} & \textbf{\textcolor{red}{0.2872}} & \textbf{\textcolor{red}{0.2277}} & \textbf{\textcolor{red}{0.4458}} & \textbf{\textcolor{red}{0.1477}} & \textbf{\textcolor{red}{0.2784}} & \textbf{\textcolor{red}{0.1650}} & \textbf{\textcolor{red}{0.2861}} & \textbf{\textcolor{red}{0.1331}} & \textbf{\textcolor{red}{0.2069}} \\
\midrule
\textbf{Improvement} & \cellcolor{pink!36}+29.76\% & \cellcolor{pink!36}+24.92\% & \cellcolor{pink!36}+14.20\% & \cellcolor{pink!36}+11.84\% & \cellcolor{pink!36}+29.47\% & \cellcolor{pink!36}+30.24\% & \cellcolor{pink!36}+25.94\% & \cellcolor{pink!36}+27.93\% & \cellcolor{pink!36}+20.25\% & \cellcolor{pink!36}+23.68\% \\
\bottomrule
\end{tabular}
}
\vspace{20pt}
\vspace*{-\baselineskip}
\centering
\caption[IMPUTATION PERFORMANCE COMPARISON]{%
  \textbf{IMPUTATION PERFORMANCE COMPARISON OF DIFFERENT METHODS \\IN TERMS OF MAE AND RMSE ON NORMALIZED DATA}
}
\label{tab:baseline_imputation}
\resizebox{\textwidth}{!}{%
\begin{tabular}{ccccccccccc}
\toprule
 & \multicolumn{2}{c}{SMS-IN} & \multicolumn{2}{c}{SMS-OUT} & \multicolumn{2}{c}{CALL-IN} & \multicolumn{2}{c}{CALL-OUT} & \multicolumn{2}{c}{INTERNET} \\
\cmidrule(lr){2-3} \cmidrule(lr){4-5} \cmidrule(lr){6-7} \cmidrule(lr){8-9} \cmidrule(lr){10-11}
Model & MAE & RMSE & MAE & RMSE & MAE & RMSE & MAE & RMSE & MAE & RMSE \\
\midrule
{\small BRITS} & \textcolor{blue}{\underline{0.0743}} & 0.1604 & 0.1412 & 0.2982 & \textcolor{blue}{\underline{0.0752}} & \textcolor{blue}{\underline{0.1641}} & \textcolor{blue}{\underline{0.0855}} & 0.1743 & 0.1331 & 0.2773 \\
{\small SAITS} & 0.0753 & \textcolor{blue}{\underline{0.1565}} & 0.1472 & 0.3026 & 0.0785 & 0.1687 & 0.0877 & \textcolor{blue}{\underline{0.1738}} & 0.1399 & 0.2797 \\
{\small LSTM} & 0.0965 & 0.1975 & 0.1489 & 0.3016 & 0.0862 & 0.1884 & 0.0989 & 0.1938 & 0.1320 & 0.2672 \\
{\small ConvLSTM} & 0.0793 & 0.1575 & \textcolor{blue}{\underline{0.1233}} & \textcolor{blue}{\underline{0.2406}} & 0.0849 & 0.1722 & 0.0965 & 0.1825 & \textcolor{blue}{\underline{0.1203}} & \textcolor{blue}{\underline{0.2225}} \\
{\small DLinear} & 0.3699 & 0.5453 & 0.4008 & 0.5731 & 0.3781 & 0.5905 & 0.3845 & 0.5832 & 0.4078 & 0.5733 \\
{\small Timesnet} & 0.0932 & 0.1912 & 0.1600 & 0.3280 & 0.0916 & 0.1898 & 0.1033 & 0.1995 & 0.1378 & 0.2857 \\
{\small STGCN} & 0.0901 & 0.1898 & 0.1342 & 0.2631 & 0.0925 & 0.1906 & 0.1012 & 0.2013 & 0.1311 & 0.2523 \\
{\small GATGPT} & 0.1289 & 0.2485 & 0.1771 & 0.3398 & 0.1372 & 0.2610 & 0.1429 & 0.2654 & 0.2113 & 0.3557 \\
{\small GPT4ST} & 0.1048 & 0.1885 & 0.1411 & 0.2886 & 0.1152 & 0.2206 & 0.1206 & 0.2294 & 0.1885 & 0.3186 \\
{\small GCNGPT} & 0.1299 & 0.2598 & 0.1705 & 0.3421 & 0.1362 & 0.2690 & 0.1420 & 0.2776 & 0.2049 & 0.3528 \\
{\small ST-LLM} & 0.1296 & 0.2865 & 0.1685 & 0.3489 & 0.1419 & 0.3010 & 0.1518 & 0.3122 & 0.2079 & 0.3629 \\
{\small ST-LLM+} & 0.1195 & 0.2407 & 0.1524 & 0.3145 & 0.1382 & 0.2768 & 0.1478 & 0.2889 & 0.1899 & 0.3250 \\
\textbf{\textcolor{red}{U-STS-LLM}} & \textbf{\textcolor{red}{0.0672}} & \textbf{\textcolor{red}{0.1372}} & \textbf{\textcolor{red}{0.0967}} & \textbf{\textcolor{red}{0.1974}} & \textbf{\textcolor{red}{0.0738}} & \textbf{\textcolor{red}{0.1570}} & \textbf{\textcolor{red}{0.0862}} & \textbf{\textcolor{red}{0.1661}} & \textbf{\textcolor{red}{0.0898}} & \textbf{\textcolor{red}{0.1637}} \\
\midrule
\textbf{Improvement} & \cellcolor{pink!36}+9.56\% & \cellcolor{pink!36}+12.33\% & \cellcolor{pink!36}+21.57\% & \cellcolor{pink!36}+17.96\% & \cellcolor{pink!36}+1.86\% & \cellcolor{pink!36}+4.33\% & \cellcolor{green!8}-0.82\% & \cellcolor{pink!36}+4.43\% & \cellcolor{pink!36}+25.35\% & \cellcolor{pink!36}+26.43\% \\
\bottomrule
\end{tabular}
}
\vspace{2pt}
\end{table*}

\begin{table}[t]
    \centering
    \caption{Robustness analysis over ten random seeds.}
    \label{tab:seed_robustness}
    \begin{tabularx}{\linewidth}{lYYY}
        \toprule
        \multirow{2}{*}{\textbf{Task}} & \multicolumn{3}{c}{\textbf{Average RMSE}} \\
        \cmidrule(lr){2-4}
        & \textbf{Mean} & \textbf{Population Std.} & \textbf{CV (\%)} \\
        \midrule
        Prediction & 0.3020 & 0.0043 & 1.4354 \\
        Imputation  & 0.1705 & 0.0054 & 3.1470 \\
        \bottomrule
    \end{tabularx}
    \par\vspace{3pt} 
    {\raggedright\footnotesize \text{Note:} The results are computed over ten random seeds: 1, 8, 11, 88, 111, 222, 333, 666, 777, 888.\par}
\end{table}

\subsection{Main Results}
\label{subsec:results}
The comprehensive performance of U-STS-LLM against all baseline models is summarized in Table \ref{tab:baseline_prediction} for the prediction task and Table \ref{tab:baseline_imputation} for the imputation task. The results demonstrate the effectiveness of the proposed unified framework.

In the long-horizon traffic forecasting task, U-STS-LLM establishes a new state-of-the-art across all five telecommunication activity types (SMS-IN, SMS-OUT, CALL-IN, CALL-OUT, INTERNET). It achieves lower errors than the specialized forecasting models, including the recent LLM-based adaptations (e.g., ST-LLM, GPT4ST) and the strongest non-LLM baseline (TimesNet). The average improvement over the best-performing baseline in terms of MAE and RMSE is 23.9\% and 23.7\%, respectively. This substantial improvement highlights the effectiveness of our spatio-temporally steered architecture in capturing complex, long-range dependencies for accurate future state estimation.

For the challenging imputation task under a high missing rate (70\%-80\%), U-STS-LLM also achieves the best overall performance. It surpasses both classic deep learning imputation methods (e.g., BRITS, SAITS) and forecasting models repurposed for reconstruction. It achieves the best performance on eight out of the ten evaluation metrics. The average improvement over the best baseline is 11.5\% for MAE and 13.1\% for RMSE. The superior imputation performance, particularly on INTERNET traffic which exhibits more complex patterns, validates the advantage of the unified learning framework. By jointly modeling the underlying data manifold for both prediction and imputation, U-STS-LLM develops a more robust and generalizable representation, enabling high-fidelity reconstruction even from severely incomplete observations.

Table~\ref{tab:seed_robustness} further verifies the robustness of U-STS-LLM across different random seeds, with the coefficient of variation of the average RMSE being only 1.44\% for traffic forecasting and 3.15\% for traffic imputation. Note that the average RMSE is obtained by averaging the RMSE values over the five traffic types.

\subsection{Ablation Studies}
\label{subsec:ablation}
To rigorously evaluate the contribution of each core design component, we conduct a systematic ablation study focusing on three critical aspects of U-STS-LLM. The results, presented in Table \ref{tab:ablation_comparison}, confirm the necessity and complementary role of each proposed module.

\begin{table}[t]
    \centering
    \caption{Ablation study comparison of the average MAE and RMSE for prediction and imputation tasks.}
    \label{tab:ablation_comparison}
    \begin{tabularx}{\linewidth}{lYYYY}  
        \toprule
        \multirow{2}{*}{\textbf{Model}} & \multicolumn{2}{c}{\textbf{Prediction}} & \multicolumn{2}{c}{\textbf{Imputation}} \\
        \cmidrule(lr){2-3} \cmidrule(lr){4-5}
         & \textbf{MAE} & \textbf{RMSE} & \textbf{MAE} & \textbf{RMSE} \\
        \midrule
        w/o STE  & 0.226 & 0.405 & 0.098 & 0.185 \\
        w/o MTL   & 0.205 & 0.356 & 0.087 & 0.177 \\
        w/o GS   & 0.212 & 0.350 & 0.125 & 0.223 \\
        w/o GE   & 0.169 & 0.305 & 0.084 & 0.165 \\
        Full Model & 0.163 & 0.301 & 0.083 & 0.164 \\
        \bottomrule
    \end{tabularx}
\end{table}

First, removing the Spatio-Temporal Embedding (STE) module leads to the most pronounced performance degradation. This module is responsible for distilling the high-dimensional raw input into a compact global context signal. Its absence deprives the LLM backbone of a structured, informative prior. The model must then infer both local patterns and global dependencies from scratch, and this markedly increases the error for both tasks.

Second, ablating the multi-task learning (MTL) objective, denoted as w/o MTL, trains the two tasks independently. This prevents the shared backbone from transferring knowledge between prediction and imputation, limiting its capacity to learn a unified spatio-temporal data manifold and reducing accuracy on both tasks.

Third, ablating the Guidance Signal (GS) generation pathway causes a significant performance drop, especially for imputation. The guidance signal provides a stable, locally-aware preliminary estimate. Without this anchor, the model is tasked with simultaneously performing coarse reconstruction and fine-grained correction, a considerably more challenging optimization problem that often converges to a suboptimal solution. The results indicate that the guidance signal effectively stabilizes the training process and facilitates residual learning.

Fourth, disabling the Graph Embedding (GE) component by removing the pre-computed functional graph also compromises model performance. This ablation eliminates the persistent structural inductive bias designed to steer the dynamic attention mechanism. Consequently, the model must rely solely on data-driven attention to discover node correlations, which proves less effective than leveraging the explicit, pre-computed functional relationships, particularly as the number of nodes scales.

The full U-STS-LLM model, integrating all three components, achieves the lowest error rates. The performance degradation observed in each ablated variant validates that the Spatio-Temporal Embedding, the Guidance Signal, and the Graph Embedding are indispensable and complementary. Together, they adapt the generic sequence modeling capacity of the LLM to the structural and statistical regularities inherent in spatio-temporal traffic data.

\subsection{Efficiency Comparison and Parameter Analysis}
\label{subsec:efficiency}

A comprehensive evaluation of model efficiency, encompassing inference latency and parameter usage, is conducted alongside the performance analysis. The results reveal the practical advantages of the proposed U-STS-LLM framework.

In terms of inference speed, U-STS-LLM demonstrates competitive latency. It is considerably faster than several representative spatio-temporal graph models and operates at a speed comparable to the most efficient baseline models.

Regarding parameter efficiency, U-STS-LLM achieves an optimal balance. While the total parameter count is moderate, a notable advantage is the substantially lower number of trainable parameters compared to other large-model-based approaches. This design considerably reduces the optimization burden and memory footprint during training. The accuracy gain is therefore achieved through an efficient architecture, rather than relying on model scale alone.

\begin{figure}[t!]
    \centering
    \includegraphics[width=0.9\linewidth]{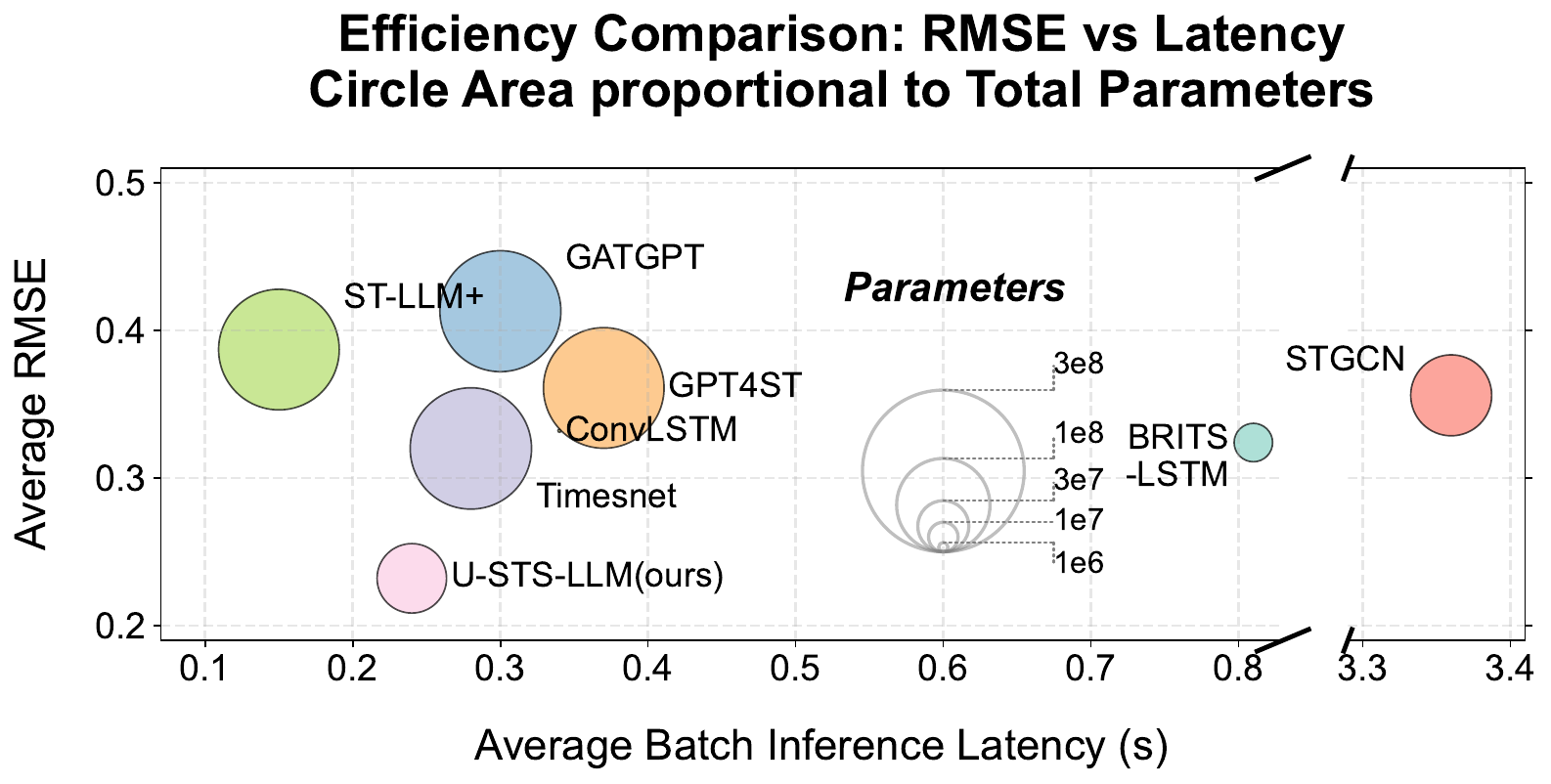}
    \caption{Efficiency comparison (circle area $\propto$ total parameters).}
    \label{fig:efficiency_total_params}
\end{figure}

\begin{figure}[t!]
    \centering
    \includegraphics[width=0.9\linewidth]{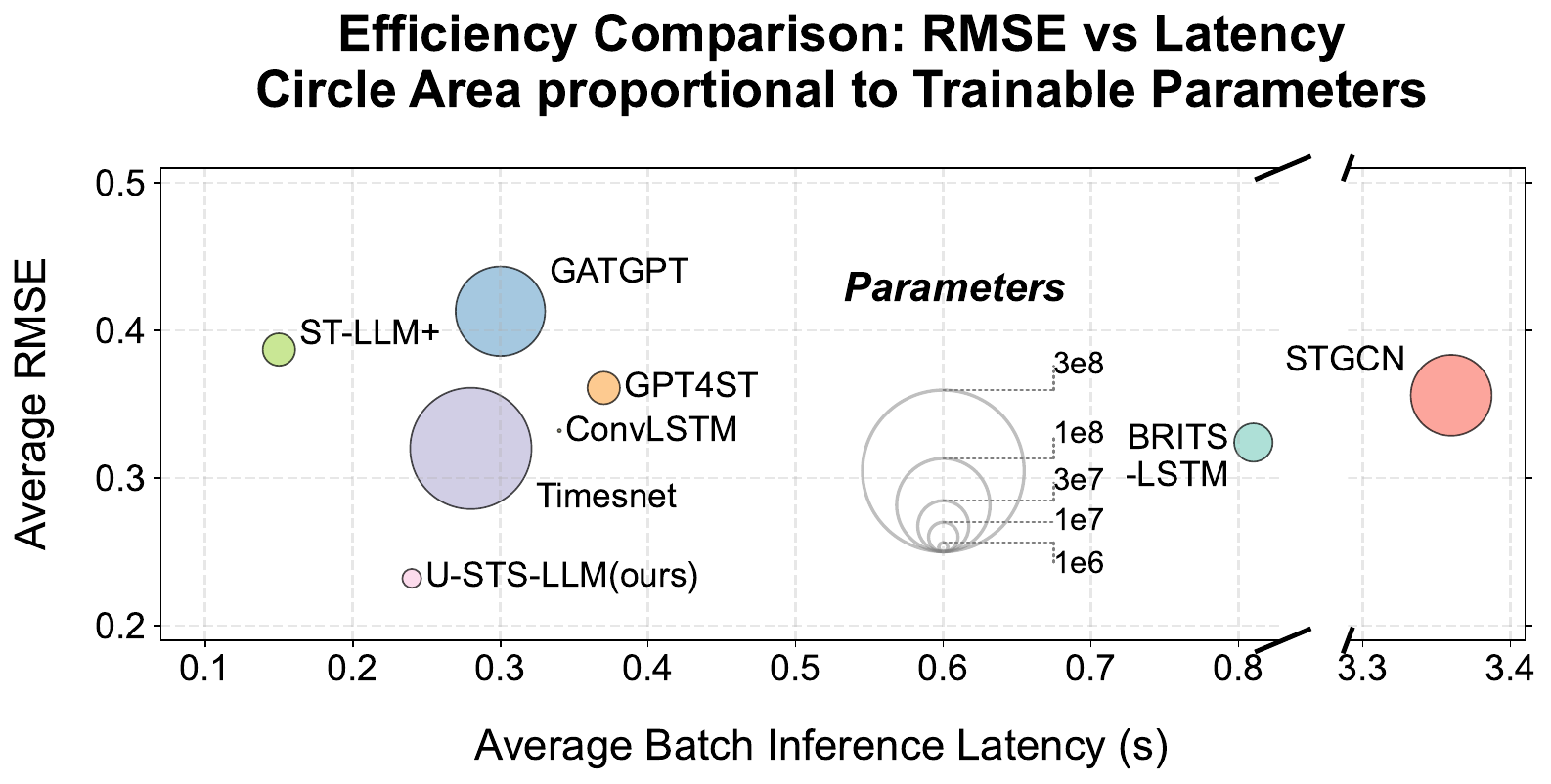}
    \caption{Efficiency comparison (circle area $\propto$ trainable parameters).}
    \label{fig:efficiency_trainable_params}
\end{figure}

Figs.~\ref{fig:efficiency_total_params} and \ref{fig:efficiency_trainable_params} position U-STS-LLM favorably in the efficiency frontier, showcasing its ability to deliver superior accuracy without incurring prohibitive computational or parametric costs. Overall, these efficiency analyses show that U-STS-LLM is an accurate and resource-efficient unified model for spatio-temporal learning.

\subsection{Analysis of Attention Patterns and Shared Layer Features Across Different Tasks}
\label{subsec:different_tasks}
\begin{figure}[hbt]
    \centering
    \begin{minipage}{0.48\linewidth}
        \centering
        \includegraphics[width=\linewidth]{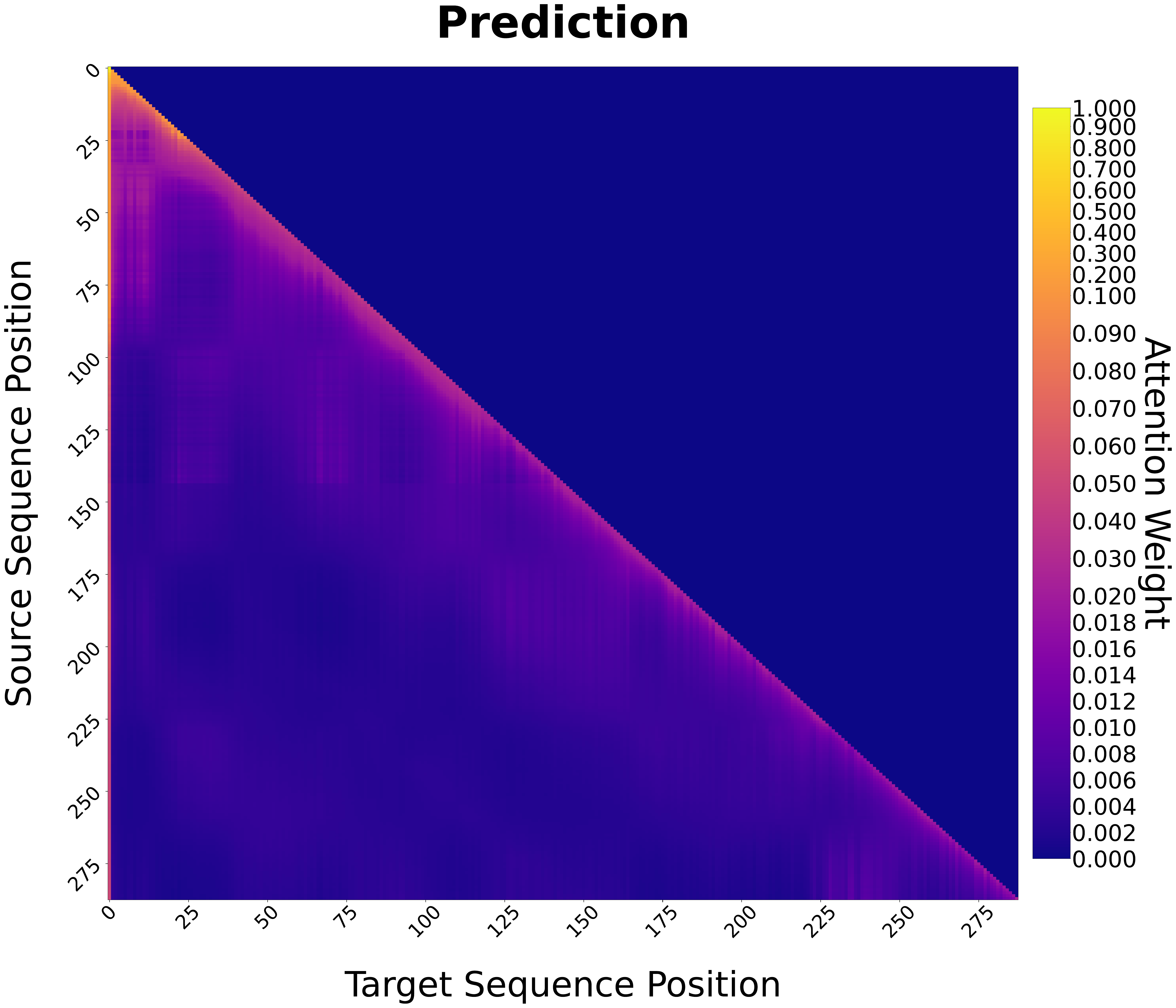}
        \caption{Average attention pattern heatmap of PFSTA across all layers in prediction tasks.}
        \label{fig:pred}
    \end{minipage}
    \hfill
    \begin{minipage}{0.48\linewidth}
        \centering
        \includegraphics[width=\linewidth]{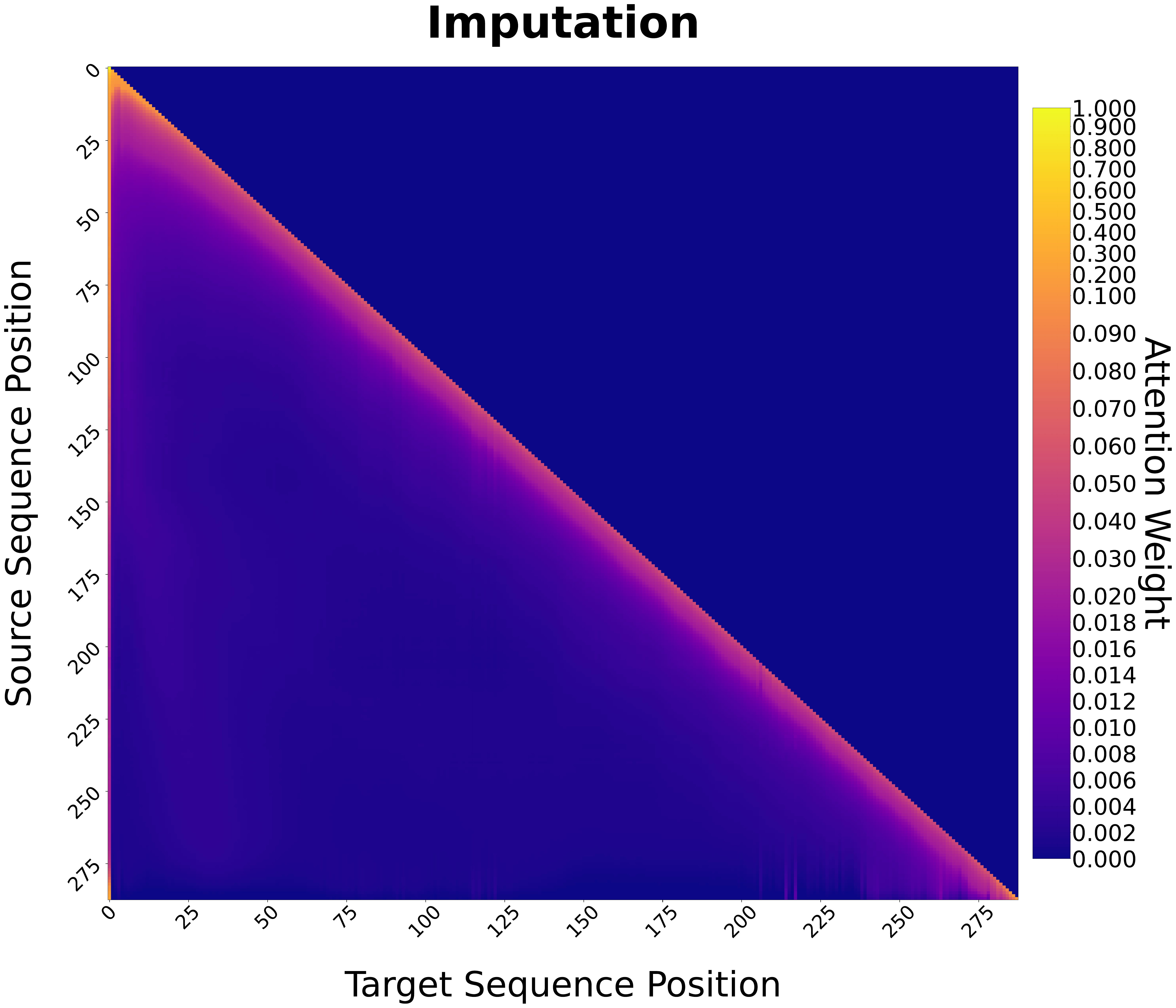}
        \caption{Average attention pattern heatmap of PFSTA across all layers in imputation tasks.}
        \label{fig:imp}
    \end{minipage}
\end{figure}

To explore how the unified multi-task architecture may support knowledge sharing and task-specific adaptation, we examine the learned attention patterns and feature representations in the shared processing layers. Attention weight matrices (Figs.~\ref{fig:pred} and \ref{fig:imp}) show both commonality and divergence across tasks. In both prediction and imputation, attention concentrates along the main diagonal, reflecting the causal, autoregressive nature of the Transformer backbone. The key difference lies in off-diagonal attention. For prediction, the model selectively attends to a few distant time steps, capturing long-term dependencies such as weekly periodic trends to support extrapolation. For imputation, attention spreads more diffusely around the diagonal and relies on local context from both past and future. This reliance is consistent with the non-causal nature of the task and allows missing values to be reconstructed smoothly while preserving continuity.

\begin{figure}[hbt]
    \centering
    \includegraphics[width=0.95\linewidth]{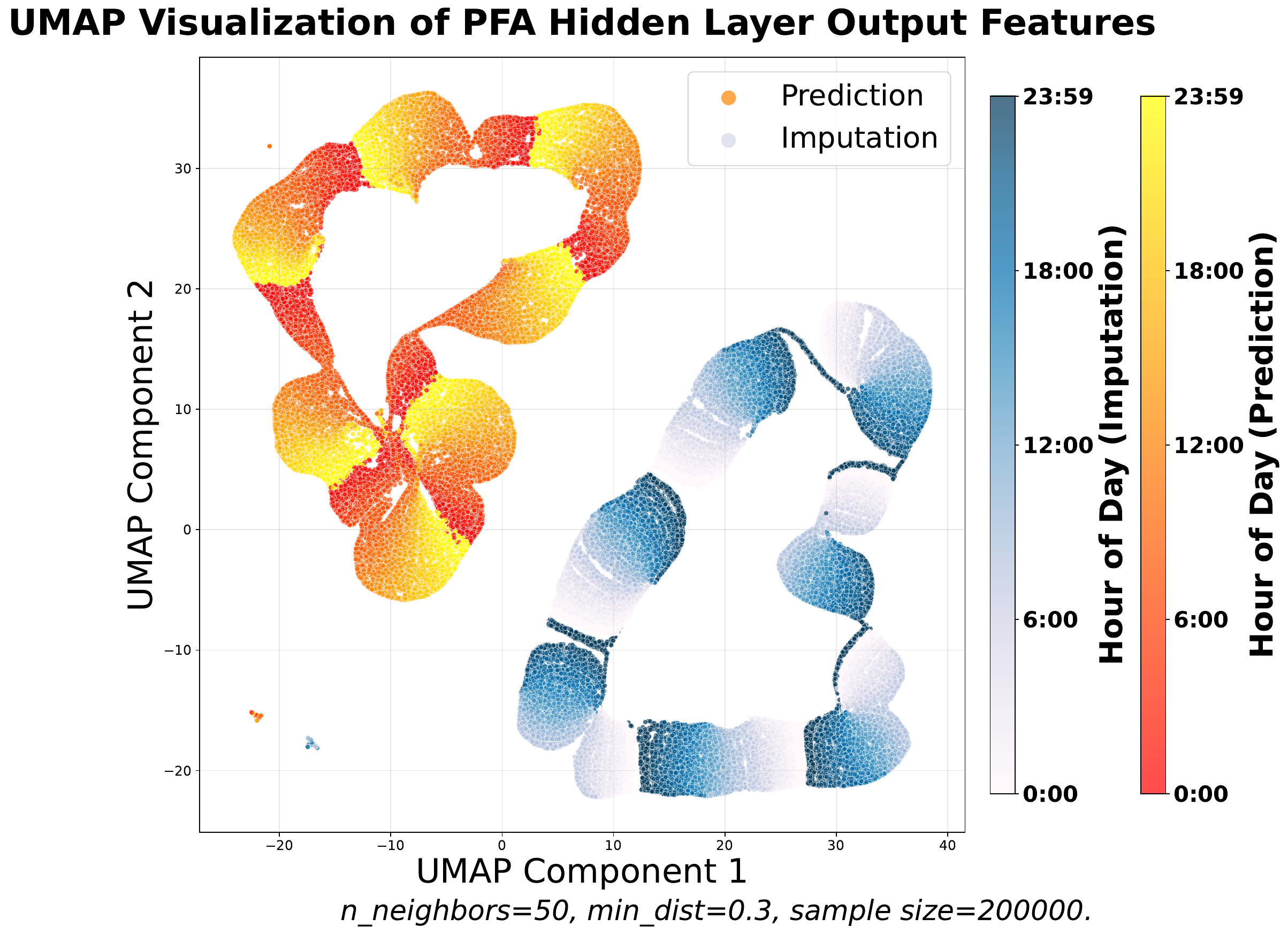}
    \caption{UMAP diagram of PFSTA hidden layer features.}
    \label{fig:umap}
\end{figure}
Further insight comes from examining the high-dimensional features learned by the initial shared PFSTA layer. A UMAP projection of these features, colored by hour-of-week, is shown in Fig.~\ref{fig:umap}. The visualization reveals two distinct, hollow ring-like structures in the reduced 2D space, one corresponding to the prediction task and the other to imputation. Both rings exhibit an identical continuous 7-period color progression, with each period spanning a day of the week and smoothly transitioning through 24 hours. This provides clear low-dimensional evidence that the shared processor has learned the fundamental weekly and daily periodic structure underlying the data, common to both tasks. The spatial separation of the rings indicates that prediction and imputation representations occupy distinct yet structurally similar subspaces, suggesting representational disentanglement that separates task-specific information while building on a shared understanding of temporal regularity.

Together, these findings reveal a coherent multi-task learning dynamic: the shared backbone extracts universal periodic regularities from spatio-temporal data, while task-conditional processing and gated fusion configure distinct pathways, namely differentiated attention and separated representation subspaces, to serve prediction (future extrapolation) and imputation (context-aware reconstruction). This synergy between shared foundational learning and task-specific specialization underpins the framework's effective and unified performance.

\subsection{Zero-Shot Performance}
\label{subsec:zero_shot}
\begin{table*}[!tbp]
\vspace*{-\baselineskip}
\centering
\caption[ZERO-SHOT PREDICTION PERFORMANCE]{%
  \textbf{ZERO-SHOT PREDICTION PERFORMANCE COMPARISON OF DIFFERENT METHODS \\IN TERMS OF MAE AND RMSE ON NORMALIZED DATA}
}
\label{tab:ZERO-SHOT_pred}
\resizebox{\textwidth}{!}{%
\begin{tabular}{ccccccccccc}
\toprule
 & \multicolumn{2}{c}{SMS-IN} & \multicolumn{2}{c}{SMS-OUT} & \multicolumn{2}{c}{CALL-IN} & \multicolumn{2}{c}{CALL-OUT} & \multicolumn{2}{c}{INTERNET} \\
\cmidrule(lr){2-3} \cmidrule(lr){4-5} \cmidrule(lr){6-7} \cmidrule(lr){8-9} \cmidrule(lr){10-11}
Model & MAE & RMSE & MAE & RMSE & MAE & RMSE & MAE & RMSE & MAE & RMSE \\
\midrule
{\small LSTM} & 0.6063 & 1.1569 & 0.4940 & 1.0242 & 0.6786 & 1.2373 & 0.5190 & 1.0963 & 0.7590 & 1.2673 \\
{\small ConvLSTM} & \textcolor{blue}{\underline{0.4490}} & 1.0213 & \textcolor{blue}{\underline{0.3908}} & 0.9541 & \textcolor{blue}{\underline{0.4118}} & 0.9853 & \textcolor{blue}{\underline{0.4521}} & 1.0437 & \textcolor{blue}{\underline{0.3952}} & \textcolor{blue}{\underline{0.8432}} \\
{\small DLinear} & 0.5990 & 1.0691 & 0.5988 & 1.0645 & 0.5639 & 1.0531 & 0.5744 & 1.0531 & 0.6601 & 1.0744 \\
{\small Timesnet} & 6.8256 & 11.8748 & 10.4643 & 21.6871 & 7.0102 & 13.1679 & 7.6870 & 14.2317 & 8.6303 & 15.9900 \\
{\small STGCN} & 0.5109 & \textcolor{blue}{\underline{0.8951}} & 0.5496 & \textcolor{blue}{\underline{0.9234}} & 0.4928 & \textcolor{blue}{\underline{0.8946}} & 0.4961 & \textcolor{blue}{\underline{0.8735}} & 0.5517 & 0.8901 \\
{\small GATGPT} & 0.7211 & 1.2421 & 1.8715 & 2.4559 & 0.5737 & 1.0094 & 0.5979 & 1.0930 & 1.2963 & 1.9081 \\
{\small GPT4ST} & 3.1559 & 4.3198 & 2.9455 & 4.0507 & 0.5014 & 1.1562 & 0.9292 & 1.4631 & 1.5407 & 2.0977 \\
{\small GCNGPT} & 1.8258 & 2.4333 & 1.5347 & 2.1419 & 1.3786 & 2.1339 & 1.4489 & 2.2392 & 0.8667 & 1.1518 \\
{\small ST-LLM} & 0.4607 & 1.0506 & 0.8937 & 1.3431 & 1.0920 & 1.4687 & 1.6524 & 3.0522 & 1.7511 & 2.4751 \\
{\small ST-LLM+} & 1.0088 & 1.1937 & 0.8608 & 1.3629 & 0.5463 & 0.9350 & 1.9627 & 3.6038 & 1.4841 & 2.1572 \\
\textbf{\textcolor{red}{U-STS-LLM}} & \textbf{\textcolor{red}{0.2020}} & \textbf{\textcolor{red}{0.5227}} & \textbf{\textcolor{red}{0.2404}} & \textbf{\textcolor{red}{0.5781}} & \textbf{\textcolor{red}{0.2146}} & \textbf{\textcolor{red}{0.5797}} & \textbf{\textcolor{red}{0.2321}} & \textbf{\textcolor{red}{0.5801}} & \textbf{\textcolor{red}{0.2352}} & \textbf{\textcolor{red}{0.5291}} \\
\midrule
\textbf{Improvement} & \cellcolor{pink!36}+55.01\% & \cellcolor{pink!36}+41.60\% & \cellcolor{pink!36}+38.49\% & \cellcolor{pink!36}+37.39\% & \cellcolor{pink!36}+47.89\% & \cellcolor{pink!36}+35.20\% & \cellcolor{pink!36}+48.66\% & \cellcolor{pink!36}+33.59\% & \cellcolor{pink!36}+40.49\% & \cellcolor{pink!36}+37.25\% \\
\bottomrule
\end{tabular}
}
\vspace{20pt}
\vspace*{-\baselineskip}
\centering
\caption[ZERO-SHOT IMPUTATION PERFORMANCE]{%
  \textbf{ZERO-SHOT IMPUTATION PERFORMANCE COMPARISON OF DIFFERENT METHODS \\IN TERMS OF MAE AND RMSE ON NORMALIZED DATA}
}
\label{tab:ZERO-SHOT_imp}
\resizebox{\textwidth}{!}{%
\begin{tabular}{ccccccccccc}
\toprule
 & \multicolumn{2}{c}{SMS-IN} & \multicolumn{2}{c}{SMS-OUT} & \multicolumn{2}{c}{CALL-IN} & \multicolumn{2}{c}{CALL-OUT} & \multicolumn{2}{c}{INTERNET} \\
\cmidrule(lr){2-3} \cmidrule(lr){4-5} \cmidrule(lr){6-7} \cmidrule(lr){8-9} \cmidrule(lr){10-11}
Model & MAE & RMSE & MAE & RMSE & MAE & RMSE & MAE & RMSE & MAE & RMSE \\
\midrule
{\small BRITS} & 0.4472 & 1.0340 & 0.4821 & 1.0222 & 0.4242 & 0.9969 & 0.4334 & 1.0090 & 0.5206 & 1.0305 \\
{\small SAITS} & 0.4413 & 1.0207 & 0.4786 & 1.0204 & 0.4128 & 0.9811 & 0.4224 & 0.9746 & 0.5638 & 1.1042 \\
{\small LSTM} & 1.0220 & 2.0773 & 0.7849 & 1.3817 & 0.5860 & 1.2164 & 0.8189 & 1.6895 & 1.3083 & 2.5986 \\
{\small ConvLSTM} & \textcolor{blue}{\underline{0.2774}} & 0.7836 & \textcolor{blue}{\underline{0.2577}} & 0.7553 & \textcolor{blue}{\underline{0.2536}} & \textcolor{blue}{\underline{0.7413}} & \textcolor{blue}{\underline{0.2786}} & 0.7903 & \textcolor{blue}{\underline{0.3142}} & \textcolor{blue}{\underline{0.7657}} \\
{\small DLinear} & 0.4021 & 0.8724 & 0.4015 & 0.8674 & 0.3812 & 0.8647 & 0.3895 & 0.8638 & 0.4490 & 0.8755 \\
{\small Timesnet} & 1.0948 & 1.8965 & 1.6523 & 3.0070 & 1.0622 & 1.8318 & 1.1324 & 1.9262 & 2.6376 & 5.5113 \\
{\small STGCN} & 0.3710 & \textcolor{blue}{\underline{0.7593}} & 0.3680 & \textcolor{blue}{\underline{0.7051}} & 0.3617 & 0.7698 & 0.3707 & \textcolor{blue}{\underline{0.7819}} & 0.4263 & 0.7793 \\
{\small GATGPT} & 1.4733 & 2.5170 & 0.6243 & 1.1155 & 1.0395 & 2.0047 & 0.8432 & 1.7145 & 0.8166 & 1.4729 \\
{\small GPT4ST} & 0.6438 & 1.0754 & 0.4745 & 0.7271 & 0.7156 & 1.6664 & 0.5007 & 0.9077 & 1.2941 & 2.4562 \\
{\small GCNGPT} & 1.1986 & 1.9907 & 1.0781 & 1.8292 & 0.7791 & 1.4338 & 1.1674 & 2.1177 & 0.9035 & 1.4113 \\
{\small ST-LLM} & 0.7288 & 1.1293 & 0.3866 & 0.7678 & 1.4400 & 2.6332 & 1.0162 & 1.7738 & 0.5071 & 0.8993 \\
{\small ST-LLM+} & 0.4186 & 0.9812 & 0.9708 & 1.4084 & 0.5896 & 0.8807 & 0.6788 & 0.9677 & 1.0959 & 2.0071 \\
\textbf{\textcolor{red}{U-STS-LLM}} & \textbf{\textcolor{red}{0.1378}} & \textbf{\textcolor{red}{0.4615}} & \textbf{\textcolor{red}{0.1490}} & \textbf{\textcolor{red}{0.4719}} & \textbf{\textcolor{red}{0.1498}} & \textbf{\textcolor{red}{0.4776}} & \textbf{\textcolor{red}{0.1666}} & \textbf{\textcolor{red}{0.4841}} & \textbf{\textcolor{red}{0.1652}} & \textbf{\textcolor{red}{0.4662}} \\
\midrule
\textbf{Improvement} & \cellcolor{pink!36}+50.32\% & \cellcolor{pink!36}+39.22\% & \cellcolor{pink!36}+42.18\% & \cellcolor{pink!36}+33.07\% & \cellcolor{pink!36}+40.93\% & \cellcolor{pink!36}+35.57\% & \cellcolor{pink!36}+40.20\% & \cellcolor{pink!36}+38.09\% & \cellcolor{pink!36}+47.42\% & \cellcolor{pink!36}+39.11\% \\
\bottomrule
\end{tabular}
}
\vspace{2pt}
\end{table*}

To rigorously assess generalization under data scarcity and domain shift, we conducted a zero-shot cross-region evaluation. U-STS-LLM and all baseline models were trained solely on the Milan telecommunications dataset and evaluated directly on the unseen Trentino dataset, without fine-tuning or exposure to the target domain. Note that, to ensure a fair comparison, U-STS-LLM directly reused the functional graph from the Milan dataset, which is valid because the model's graph-based operations do not depend on any city-specific node semantics. This zero-shot protocol simulates deployment in a new city or network topology with potentially different spatio-temporal patterns, providing a strict test of the model’s ability to learn transferable principles of spatio-temporal dynamics rather than memorizing location-specific idiosyncrasies.

The zero-shot prediction and imputation results are summarized in Table \ref{tab:ZERO-SHOT_pred} and Table \ref{tab:ZERO-SHOT_imp}, respectively. The performance gap observed in the standard in-domain evaluation widens dramatically in this cross-domain setting, unequivocally highlighting the superior generalization prowess of U-STS-LLM. For the long-horizon forecasting task, U-STS-LLM achieves a decisive and consistent advantage across all five telecommunication activity types. It significantly outperforms all specialized baselines, including recent LLM adaptations (e.g., ST-LLM, GPT4ST) and robust spatio-temporal models (e.g., STGCN, ConvLSTM). The average improvements in MAE and RMSE both exceed 37\%, a margin substantially larger than that observed in the in-domain test. This indicates that while many models suffer from catastrophic performance degradation when facing a shifted data distribution, U-STS-LLM maintains robust and accurate predictive capability. Notably, some powerful in-domain models like TimesNet exhibit extreme performance collapse, suggesting severe overfitting to the source domain's specific periodic patterns and scale. Conversely, the success of U-STS-LLM underscores the effectiveness of its core design: the functional graph based on traffic profile similarity provides a portable structural prior that transcends specific geographical coordinates; the conditional input scaling enhances robustness to distributional shifts in data magnitude; and the unified multi-task optimization compels the model to learn a holistic, disentangled representation of the underlying data manifold that is invariant to superficial domain characteristics.

Similar advantages are observed for zero-shot imputation under a high missing rate. U-STS-LLM achieves the best performance on all evaluation metrics, surpassing dedicated imputation methods like BRITS and SAITS, which are themselves designed for robustness. The average improvement is approximately 44\% for MAE and 37\% for RMSE. This is a particularly compelling result because imputation under a high missing rate heavily relies on a correct inductive bias of the data's joint spatio-temporal distribution. The fact that U-STS-LLM, trained only on Milan's data, can accurately reconstruct large missing blocks in Trentino's data strongly suggests that the model has indeed learned a generalizable prior for how traffic states evolve and correlate across space and time. The joint training objective, which forces the model to solve both forward prediction (extrapolation) and inverse imputation (interpolation), appears to foster a more complete and transferable understanding of the system dynamics than models trained for a single objective.

The observed zero-shot performance supports the main claim of this work. The proposed architecture, which steers a pre-trained LLM with explicit, learnable, and portable spatio-temporal biases, mitigates the generalization limitations of specialized models. The Dynamic Spatio-Temporal Attention Bias Generator provides a relational prior that remains effective beyond the training domain. The partially frozen backbone with LoRA tuning preserves the LLM's general sequence reasoning while efficiently specializing its attention mechanism. The unified multi-task optimization learns a shared representation that captures essential spatio-temporal dynamics. Together, these mechanisms enable U-STS-LLM to maintain strong standard performance and achieve effective zero-shot transfer, offering a practical approach for building foundation models on structured, non-linguistic data.

\section{Conclusion}
\label{sec:conclusion}
This paper presented U-STS-LLM, a unified framework that overcomes task decoupling, spatio-temporal decoupling, and error-prone serial chaining, the three structural limitations of prior spatio-temporal traffic models. By steering a partially frozen spatio-temporally attentive (PFSTA) LLM backbone with a dynamic spatio-temporal attention bias, and fusing its deep representation with a stable local estimate via Gated Adaptive Fusion, the model confines the LLM to residual correction under a unified multi-task objective. This design avoids the accumulation of errors and latency inherent in cascaded pipelines, while remaining parameter-efficient through LoRA-based partial freezing and portable across deployments without relying on costly physical graphs.

Extensive experiments confirm that U-STS-LLM attains state-of-the-art accuracy in both long-horizon forecasting and high-missing-rate imputation, with low trainable-parameter overhead and strong zero-shot generalization to an unseen network region. The attention- and representation-level analysis further corroborates that the shared backbone captures universal daily/weekly periodic structure, while task-conditional pathways achieve representational disentanglement for extrapolation versus reconstruction.

Despite these results, this study is constrained by limited training time, GPU resources, and the scarcity of publicly available cellular traffic datasets. Consequently, we have not yet evaluated the model's scalability to larger node-aggregation scales or longer time horizons, nor tested its zero-shot transfer behavior on other spatio-temporal traffic datasets with markedly different topologies. Exploring these regimes, together with extending the steering paradigm to multi-modal urban contexts and real-time adaptive deployment, remains a promising direction for future work.



\bibliographystyle{IEEEtran}  
\bibliography{IEEEabrv, myrefs} 



\vfill
\end{document}